\documentclass[a4paper,10pt,conference]{IEEEtran}
\usepackage{etoolbox}
\makeatletter
\patchcmd{\@makecaption}
  {\scshape}
  {}
  {}
  {}
\makeatother
\ifCLASSINFOpdf
\else
\fi
\usepackage{amsmath}
\usepackage{amsfonts}
\usepackage{graphicx}
\usepackage{tikz}
\usepackage{pgfplots}
\begin{document}
%
\title{Conditional Transfer with Dense Residual Attention: Synthesizing traffic signs from street-view imagery}



%
\author{\IEEEauthorblockN{
\IEEEauthorrefmark{1}Clint Sebastian\IEEEauthorrefmark{2},\IEEEauthorrefmark{1}Ries Uittenbogaard\IEEEauthorrefmark{4},
Julien Vijverberg \IEEEauthorrefmark{3},
Bas Boom\IEEEauthorrefmark{3},
and Peter H.N. de With\IEEEauthorrefmark{2}}
\IEEEauthorblockA{\IEEEauthorrefmark{4} Department of Mechanical Engineering, Delft University of Technology, Delft, The Netherlands
}
\IEEEauthorblockA{\IEEEauthorrefmark{2}Department of Electrical Engineering,
Eindhoven University of Technology, 
Eindhoven, The Netherlands}
\IEEEauthorblockA{\IEEEauthorrefmark{3} Cyclomedia Technology B.V, Zaltbommel, The Netherlands \\
Email: m.c.uittenbogaard@student.tudelft.nl, \{c.sebastian, p.h.n.de.with\}@tue.nl,} 
\{jvijverberg, bboom\}@cyclomedia.com \\
\IEEEauthorrefmark{1} \textit{denotes equal contribution}
}



\maketitle
\begin{abstract}
Object detection and classification of traffic signs in street-view imagery is an essential element for asset management, map making and autonomous driving. However, some traffic signs occur rarely and consequently, they are difficult to recognize automatically. To improve the detection and classification rates, we propose to generate images of traffic signs, which are then used to train a detector/classifier. In this research, we present an end-to-end framework that generates a realistic image of a traffic sign from a given image of a traffic sign and a pictogram of the target class. We propose a residual attention mechanism with dense concatenation called Dense Residual Attention, that preserves the background information while transferring the object information. We also propose to utilize multi-scale discriminators, so that the smaller scales of the output guide the higher resolution output. We have performed detection and classification tests across a large number of traffic sign classes, by training the detector using the combination of real and generated data. The newly trained model reduces the number of false positives by 1.2~-~1.5\% at 99\% recall in the detection tests and an absolute improvement of 4.65\% (top-1 accuracy) in the classification tests.
\end{abstract}


%
\IEEEpeerreviewmaketitle

\section{Introduction}

Detection and classification of traffic signs in street-view imagery is vital for public object maintenance, map making and autonomous driving. This is particularly challenging when certain classes or categories of objects are scarce. Challenging cases occur in the automated detection and classification of traffic signs at a country-wide level on high-resolution street-view imagery. Manual efforts to find traffic signs in millions of high-resolution images is cumbersome and the detection/classification algorithm fails if it is not trained with the class-specific data or when traffic signs rarely occur. A possible approach to alleviate this problem is to generate realistic data using generative modeling and expand the training sets that have low amount of data or low recognition scores. However, generation of photo-realistic samples of traffic signs is difficult due to large variations in pose, lighting conditions and varying background. 

Recent developments in deep learning can be applied to modeling of image-to-image translation problems. Generative Adversarial Network (GAN) is a class of  deep learning algorithms that is used for generative modeling \cite{goodfellow2014generative}. GANs formulate the generative modeling problem as zero-sum game between two networks. A GAN consists of a generator network that produces samples from a given input sample or noise and a discriminator network that tries to distinguish if the generated sample is from the real or fake data distribution. Although Convolutional Neural Networks (CNNs) may be used to perform image-to-image translations, many of them apply a stochastic approximation to minimize an objective function and require paired data \cite{zhang2016colorful}\cite{johnson2016perceptual}. Alternatively, GANs try to achieve Nash equilibrium by generating a distribution that is close to the empirical one.

Conditional variants of GANs have recently produced impressive results in image-to-image translation applications \cite{kim2017learning}\cite{mirza2014conditional}\cite{isola2016image}\cite{zhu2017unpaired}\cite{wang2017highres}. A conditional GAN tries to map a Domain~\textit{A} to another Domain~\textit{B}, instead of using a noise vector as input. It learns to translate an underlying relationship between the Domains~\textit{A} and~\textit{B} without explicitly pairing inputs and outputs.
To have a better consistency for the mapping, most Conditional GANs also reconstruct the output back to the input \cite{kim2017learning}\cite{zhu2017unpaired}. Apart from the mapping from Domain~\textit{A} to~\textit{B}, Domain~\textit{B} is also reconstructed back to~\textit{A}. The loss from the inverse mapping is also added to the objective function as a regularizer during training. Image analogy problems can be modeled using a Conditional GAN by providing an auxiliary piece of information~\cite{Jetchev_2017_ICCV_Workshops}. The auxiliary information could be text, image or other data modalities. For example, an input image of a traffic sign can be paired with a pictogram to obtain an output image of a new traffic sign.

In this paper, we thus explore a conditional GAN for traffic sign generation, while using an auxiliary information of a pictogram. Specifically, we address the problem of retaining the original background while altering only the object information. In more detail, we propose a conditional GAN with Dense Residual Attention. To further improve the texturing and details of the generated traffic sign, we use multi-scale discriminators. We reduce the dependency of the mask generation process that is used in recent work~\cite{Jetchev_2017_ICCV_Workshops}. Our method produces perceptually acceptable results without the implicit generation of a mask. However, we are able to obtain better results with a weak supervision on the traffic signs that have a complex pose. Finally, we improve classification and detection rates of rare traffic signs in high-resolution street-view imagery. 

\section{Related Work}
Unsupervised generative modeling using GANs has achieved state-of-the-art results in recent years. Many works have produced perceptually realistic images for problems such as image-to-image translation~\cite{kim2017learning}\cite{mirza2014conditional}\cite{isola2016image}\cite{zhu2017unpaired}\cite{wang2017highres}, image in-painting \cite{pathak2016context}, super-resolution \cite{johnson2016perceptual}\cite{ledig2016photo}\cite{Tai_2017_CVPR} and conditional analogy~\cite{Jetchev_2017_ICCV_Workshops}. Deep Convolution GANs (DCGANs) introduced the convolutional architecture GANs that improve visual quality of generated images \cite{radford2015unsupervised}. Recently, Wasserstein GANs (WGANs) introduced an objective function that improves model stability and provides a meaningful measure of convergence~\cite{arjovsky2017wasserstein}. However, they require the discriminator (also known as critic) to lie within the space of 1-Lipschitz functions. To enforce the Lipschitz constraint, the weights are clipped to a compact space. To circumvent weight clipping, improved WGAN have proposed to add a gradient penalty term to the WGAN training objective. This gradient penalty term is the product of the penalty coefficient term and gradient norm of the critic's output. This approach results in a better performance and stability~\cite{gulrajani2017improved}.

Recently, GANs have gained popularity in image-to-image translation problems. GANs have been applied in a conditional setting to generate text, labels and images.  Image-conditional models such as ``pix2pix", use a Conditional Adversarial Network to learn paired mappings between domains~\cite{isola2016image}. CycleGAN performs an unpaired domain transfer by adding a cycle consistency loss to the training objective~\cite{zhu2017unpaired}. Similarly, DiscoGAN also proposes an unpaired domain transfer where two reconstruction losses are added to the objective function~\cite{kim2017learning}. Conditional Analogy GAN (CAGAN) propose to swap clothing articles on people~\cite{Jetchev_2017_ICCV_Workshops}. Given a human model and a given fashion article, it produces a human model wearing the fashion article. The generator of the CAGAN architecture generates both the image and an implicit segmentation mask. The final output image is a convex combination of both the generated image and input image. However, in scenarios where the background is complex, the generation of an implicit mask becomes a challenging task. This is addressed in~\cite{Jetchev_2017_ICCV_Workshops}.

Attention mechanisms are necessary for proper guidance of feature propagation. Many efforts have been made towards incorporating an attention mechanism in deep neural networks~\cite{NIPS2010_4089}\cite{kim2016multimodal}\cite{zheng2017learning}. Attention mechanisms have been widely adopted in Recurrent Neural Networks (RNNs) and Long Short Term Memory (LSTM)~\cite{hochreiter1997long} for modeling sequential tasks. Residual Attention Networks propose an attention mechanism by stacking multiple attention modules for various classification tasks~\cite{he2016deep}\cite{Wang_2017_CVPR}. Each attention module consist of a mask and a trunk branch. The trunk branch performs feature processing using residual units and the mask branch uses a bottom-up top-down structure to guide the feature learning. Densely Connected Convolutional Network (DenseNet) proposes an architecture where every layer is connected to all higher layers~\cite{Huang_2017_CVPR}. This type of connectivity enables reusable features and provides high parameter efficiency, as low-dimensional feature maps are reused recursively. Concurrent to our work, a multi-scale discriminator approach has been explored in~\cite{wang2017highres} for image synthesis from semantic labels. In our method, we use a multi-scale discriminator to learn the finer details of the pictogram and the global features such as the pose and lighting condition of the traffic sign. The methodology is described in the following section.

\section{Method} 

The task of transferring a pictogram to a given traffic sign can be formulated as follows. Given an image \boldmath{$x_i$} of a traffic sign and pictogram \boldmath{$p$} of the target class, the generator network~\textit{G} tries to produce a traffic sign image of the target class \boldmath{${{x}_i^p}$}. The discriminator network \textit{D} distinguishes between \boldmath{${{x}_i^p}$} and \boldmath{${{x}_j}$}, where \boldmath{${{x}_j}$} is sampled from the real data distribution. 

For training, we use the improved WGAN objective with cycle consistency loss at multiple scales~\cite{gulrajani2017improved}\cite{zhu2017unpaired}. The final training objective $\mathcal{L^{\textit{s}}}$ at a given scale \textit{s} is expressed as:
\begin{equation}
\mathcal{L^{\textit{s}}} = 
\underset{\textit{G}}{\textnormal{min }} \underset{\textit{D}^s}{\textnormal{max }} \mathcal{L^{\textit{s}}_{\textnormal{WGAN-GP}}} (\textit{G}, \textit{D}^s) + \mathcal{L^{\textit{s}}_{\textnormal{cyc}}} (\textit{G}),
\end{equation}
where  $\mathcal{L_{\textnormal{WGAN-GP}}} (\textit{G}, \textit{D})$ is the WGAN loss function~$\mathcal{L_{\textnormal{WGAN}}}$ with gradient penalty and $\mathcal{L_{\textnormal{cyc}}} (\textit{G})$ is the cycle loss. The training objective  $\mathcal{L_{\textnormal{WGAN}}}$ is expressed as the difference of the expected values of the fake and real outputs from the discriminator~\textit{D}. The WGAN adversarial loss with gradient penalty for our problem now becomes: 
\begin{equation}
	\begin{aligned}
		\mathcal{L^{\textit{s}}_{\textnormal{WGAN-GP}}} (\textit{G}, \textit{D}^s)  = 
		\underset{{x}_i^p \sim \mathbb{P}_{g}} 
		{\mathbb{E}}[\textit{D}^s({x}_i^p)] - 
		\underset{x_j^p \sim \mathbb{P}_{r}} 
		{\mathbb{E}}[\textit{D}^s({x}_j)] \\
		 + \lambda
		\underset{\hat{x}_i \sim \mathbb{P}_{\hat{x}_i}} 
		{\mathbb{E}}
        [({\left \| {\nabla}_{\hat{x}_i} \textit{D}^s(\hat{x_i}) \right \|}_2 - 1)^2].
	\end{aligned}
\end{equation}
Here, the output from the generator \textit{G}(\boldmath{${{x}_i}$}$| p$) $\approx$  \boldmath{${{x}_i^p}$} and $\hat{x}_i$ is a sampling from the distribution $\mathbb{P}_{\hat{x}_i}$. The sample $\hat{x}_i$ is an interpolation between a pair of samples from the generated and real distributions $\mathbb{P}_{g}$  and $\mathbb{P}_{r}$. The gradient penalty coefficient term $\lambda$ is set to 10 for all the experiments in this research, which was adequate for our work and in accordance with \cite{gulrajani2017improved}. To push the norm of the gradient towards unity, the gradient penalty is applied. For the cycle consistency loss, we use the same generator network to map multiple classes in a given category. Hence, cycle loss for our objective is defined as: 
\begin{equation}
\mathcal{L^{\textit{s}}_{\textnormal{cyc}}} (\textit{G}) = 
\underset{{x}_i^{p_a} \sim \mathbb{P}_{g}} 
{\mathbb{E}}
{\left \| {x}_i^{p_{a}} - \textit{G}(\textit{G}({{x}_i^{p_{a}}}| p_{b})|p_{a}) \right \|}_2,
\end{equation}
\begin{figure*}[t]
\begin{center}
   \includegraphics[width=1.0\linewidth]{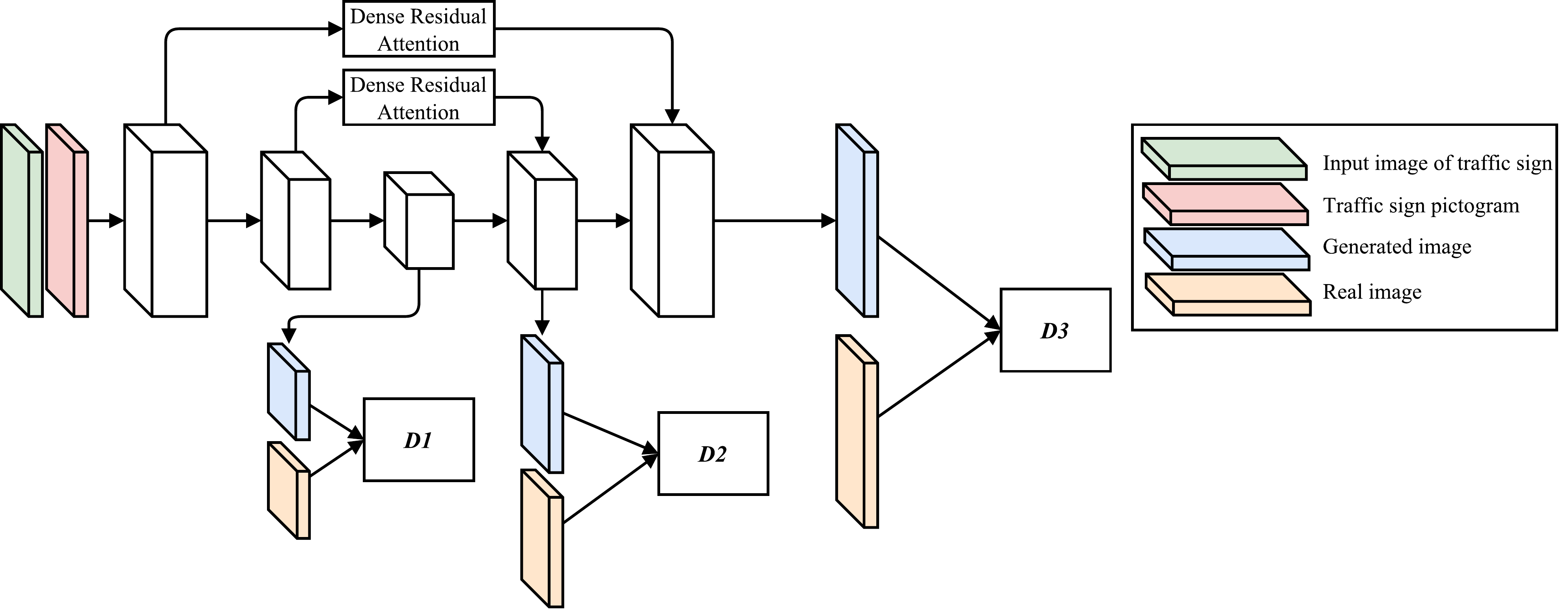}
\end{center}
   \caption{Proposed network with multiple discriminators ($D1, D2, D3$). The network uses a Dense Residual Attention module and a discriminator at each scale. The Dense Residual Attention module receives the input from the feature maps $F_e$ at the encoder side and outputs $F^{out}_{d}$. The output image generated from the auxiliary branch is supplied to the discriminator at a given scale. When the mask is applied, an element-wise product of the mask and the generated image is fed to the discriminator. The cycle loss is also obtained using the same generator network. Cycle loss not shown in figure. (Best viewed in color). }
\label{fig:dranet_refined}
\end{figure*} 
where $p_a$ and $p_b$ are pictograms of different traffic signs. The samples ${x}_i^{p_a}$ and ${x}_i^{p_b}$ are traffic signs conditioned on $p_a$ and $p_b$. Note that $\textit{G}$(${x}_i^{p_a}$ $| p_b$)~$\approx$~${x}_i^{p_b}$ and $\textit{G}$(${x}_i^{p_b}$ $| p_a$)~$\approx$~${x}_i^{p_a}$. By transitive relation, $\textit{G}(\textit{G}({{x}_i^{p_{a}}}| p_{b})|p_{a})$ $\approx$ ${x}_i^{p_{a}}$. Therefore we also minimize the $L_2$ distance between the input and reconstructed examples. An overview of the proposed network is presented in Figure~\ref{fig:dranet_refined}. The generator has an encoder-decoder structure with a residual attention mechanism and dense connectivity at each scale. To  discriminate between the real and generated samples at multiple scales, a discriminator is applied at each scale. The specifics of our contributions are addressed in the following subsections.

\subsection{Dense connectivity}
To enhance the feature propagation, we apply a dense connection between the encoder and decoder of the generator network. The dense connectivity is achieved by concatenating feature maps of the same size from the encoder to the decoder. Since the convolution over concatenated feature tensors is computationally expensive, we apply a 1 $\times$ 1 convolution across channels to reduce dimensionality \cite{nin2014}.  The output feature tensor $F^{c}_d$ at the decoder side is given as 

\begin{equation}
F^{c}_{d} = {[F_d, F_e]}_{1 \times 1},  
\end{equation}

where $F_e$ and $F_d$ are the feature maps at a given scale in the encoder and decoder. The expression $[.,.]_{1 \times 1}$ denotes the $1 \times 1$ convolution followed by the concatention operation $[.,.]$. The $1 \times 1$ convolution reduces the dimensionality of $F^{c}_d$ to the same size as $F_e$.
\begin{figure}[h]
\begin{center}
   \includegraphics[width=1.0\linewidth]{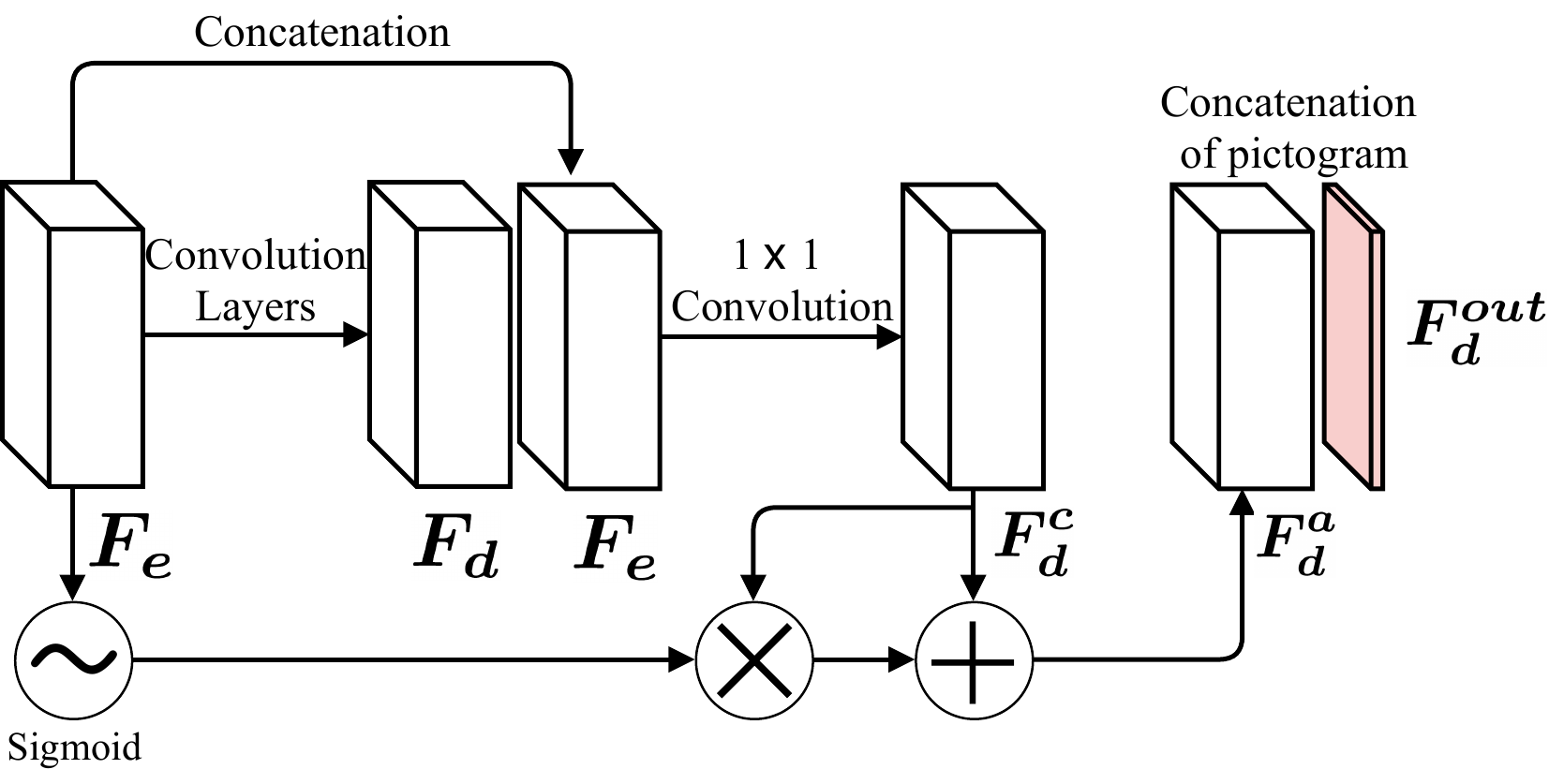}
\end{center}
   \caption{ Dense Residual Attention module followed by concatenation of the pictogram. Feature maps from the encoder $F_e$ are concatenated with $F_d$, followed by 1$\times$1 convolution to produce $F_d^{c}$. The Hadamard product of $F_e$ and $F_d^{c}$ followed by the addition of $F_d^{c}$, results in $F_d^{a}$. The output obtained from the Dense Residual Attention module $F_d^{a}$ with the pictogram $p$ represents~$F^{out}_d$. }
\label{fig:DRA}
\end{figure} 
\subsection{Attention mechanism through residual connections}
The attention mechanism is necessary for learning the relevant features that must be passed to the subsequent layers. To retain the background of the input image, the features closer to the input should be preserved. The encoder side of the generator has features closer to the input and hence the information from the encoder is transferred to the decoder through a residual attention mechanism. Our method is similar to the approach proposed in \cite{Wang_2017_CVPR}. However, we do not require a mask or trunk branch for feature processing, instead we couple the output feature maps from encoder and decoder. The proposed attention mechanism does not require any additional trainable parameters. The updated feature maps $F^a_d$ after the attention mechanism is expressed as:

\begin{equation}
F^a_d =  F^{c}_d + \sigma (F_e) \odot F^{c}_d,  
\end{equation}

where $\sigma$ denotes the sigmoid activation and $\odot$ denotes the Hadamard product. At a given scale, we update the feature tensor in the decoder by the element-wise product of  $\sigma ( F_e)$ and $F^c_d$ followed by the addition of $F^c_d$. We have also found that concatenating the pictogram $p$ of the desired traffic sign at each scale improves the performance. The output tensor $F^{out}_d$ at the decoder side is [$F^a_d$, $p$]. At larger scales, the concatenation of the traffic sign pictogram preserves the finer details of the target traffic sign.
We refer to the combination of residual attention mechanism and dense connectivity as \textit{Dense Residual Attention}. The Dense Residual Attention module is illustrated in Figure \ref{fig:DRA}.
\subsection{Multi-scale discriminators}
To understand both the global and finer details, we train a discriminator at each scale. Concurrent to our work, multi-scale discriminators are used in \cite{wang2017highres}. However, our generator has auxiliary branches that generate an output image at each scale. We use multiple optimizers which are optimized jointly. We generate outputs after up-sampling the outputs from the previous layer by a factor of two. The generated image at a given scale is fed to the corresponding discriminator. To reduce computational cost, the generated image at a smaller scale use a discriminator of smaller depth. We use three discriminator networks ($D1, D2, D3$) which receive the input from the auxiliary branches of the generator. The discriminator at the smallest scale attends to global features such as lighting condition and pose of the traffic sign, whereas the concatenated pictogram $p$ along with the discriminator at the largest scale captures the finer details of the traffic sign. Multi-scale discriminators simplify the transition going from the coarsest to the finest scale by retaining the global features.  
\begin{figure}[h]
\centering
   \includegraphics[width=1.0\linewidth]{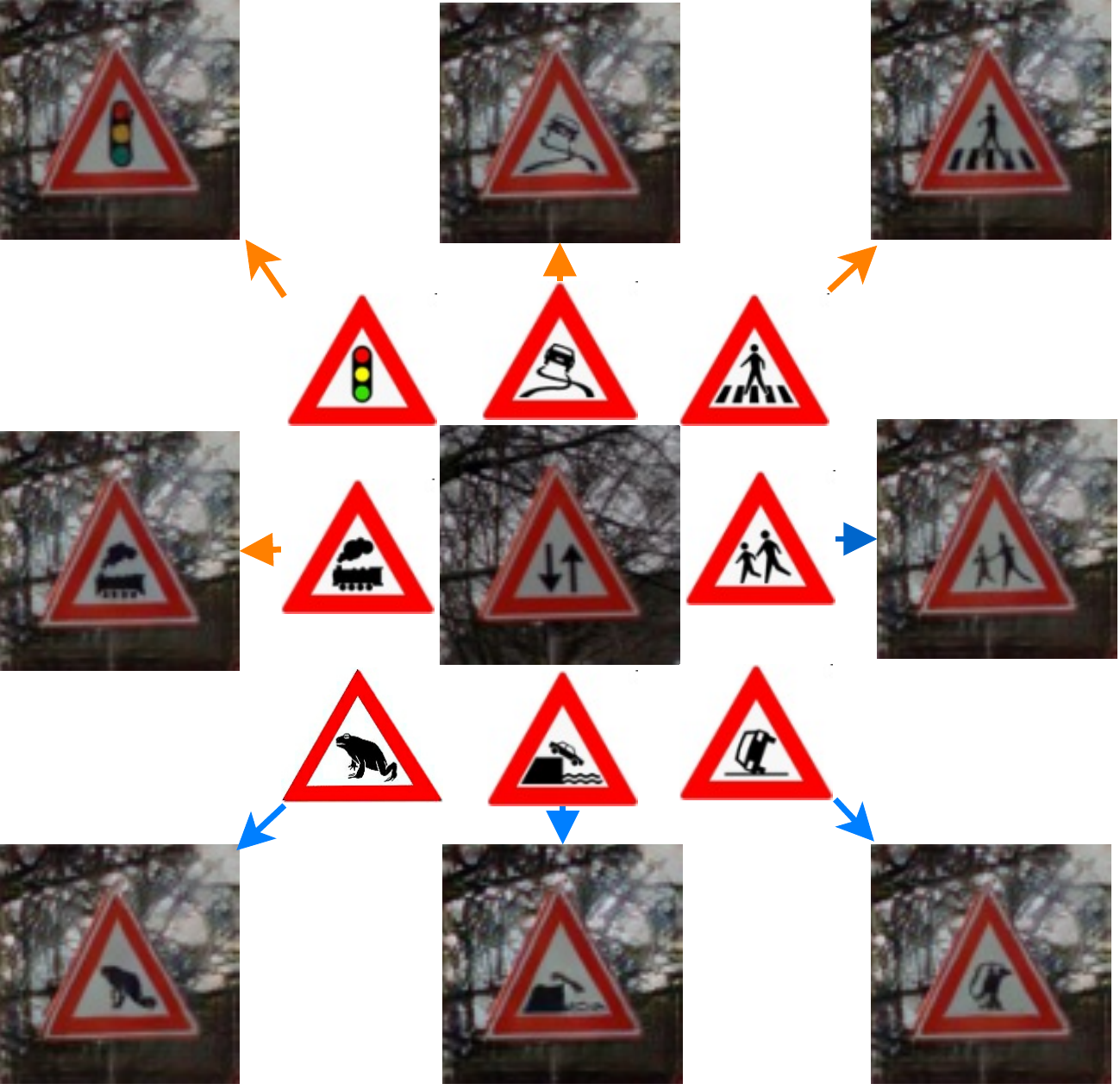}
   \caption{ Given an image of a traffic sign (center) and a pictogram, the trained model generates an image of a new traffic sign. Orange and blue arrows represent classes (not examples) inside and outside the training set.}
\label{fig:results}
\end{figure} 
\subsection{Mask for weak supervision}
In previous work~\cite{Jetchev_2017_ICCV_Workshops}, an implicit mask is generated to attend to the desired object. A convex combination of the input and the generated image produces the output image. This implicit mask generation is a tedious task when the desired object has varying light conditions and a cluttered background. With the methods described in the previous subsections, we have obtained perceptually appealing results when the pose of the traffic sign is not too skewed. However, we have found it beneficial to apply a mask to improve the performance. We use a rectangular bounding box on the desired object that intensity set to unity. The region of the mask around the bounding box has an intensity range within the unit interval, which changes during training starting from one. The final output that is supplied to the discriminator is the element-wise product of the mask and the generated image. The provided mask is only required during training and is not used during testing. 

\begin{figure}[t]
\centering
   \includegraphics[width=1.0\linewidth]{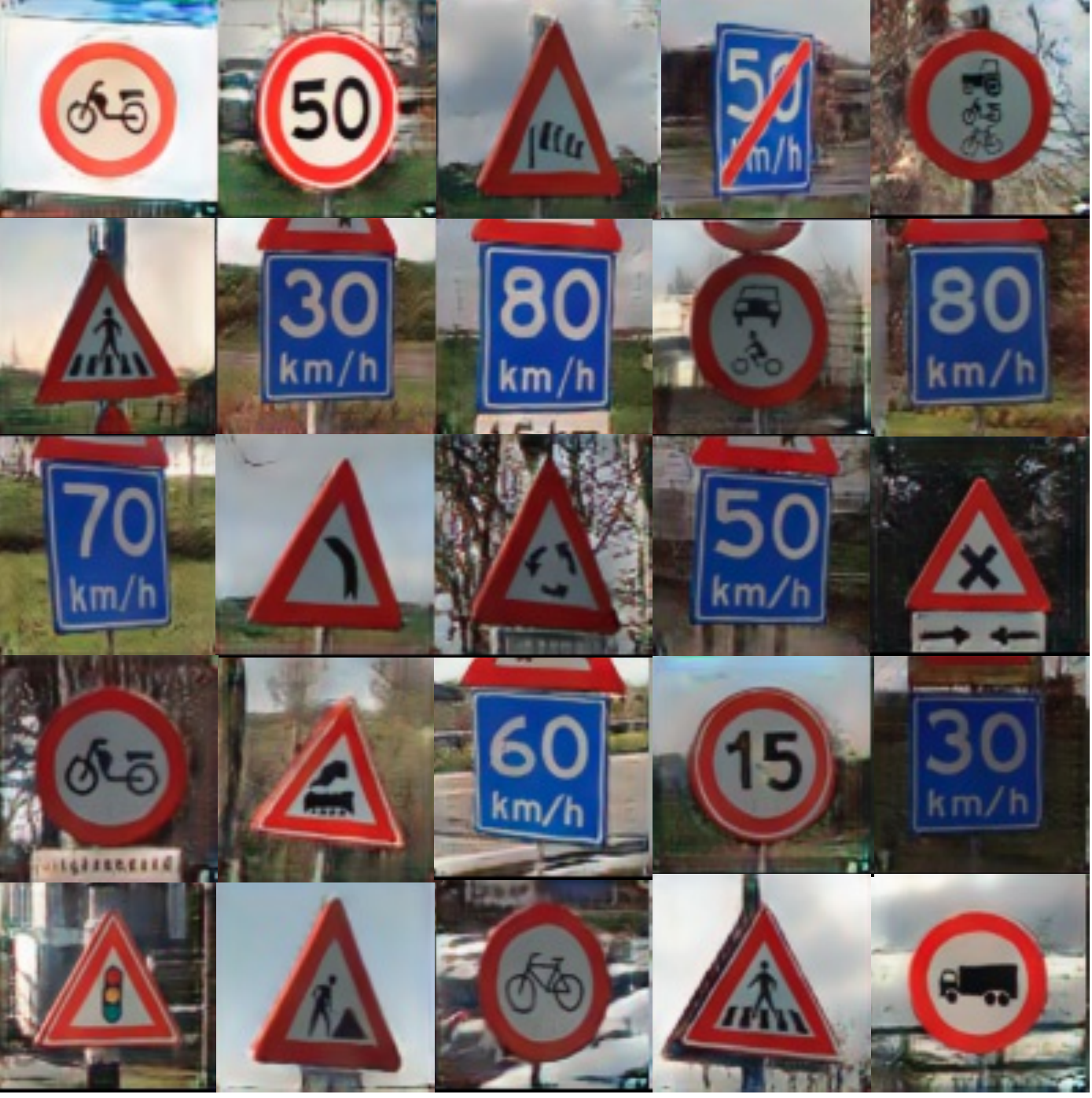}
   \caption{ Examples of generated traffic signs using our method.}
\label{fig:results_multi}
\end{figure} 
\section{Experiments}
\subsection{Dataset}
We have obtained the images of Dutch traffic signs within high-resolution street-view images from the company Cyclomedia and the pictograms from \cite{ts_dutch}. Out of a total of 313~classes, we select images of traffic signs from 55~classes that have a low amount of data and low recognition rates. We broadly partition the images into three categories, based on the appearance of the traffic signs as white triangles, white circles and blue rectangles. Each of the image and the pictogram has a resolution of 80 $\times$ 80 pixels.

\begin{table*}[t]
\begin{center}
\caption{Classification performance of three categories of traffic signs that consist of 55 classes. Each class is approximately expanded by 300 examples.}
  \begin{tabular}{ | c | c | c | c | c | c | c | }
    \hline
    \multicolumn{2}{|c|} {Category information}& \multicolumn{2}{|c|} {Amount of training data} & \multicolumn{3}{|c|}{Classification performance (Top-1 score)} \\ \hline
    Category & Number of classes & Real training data & Generated training data & Real data & Real + Generated data & Difference \\ \hline
    White triangles & 26 & 6498 & 7828 & 50.0\% & 55.3 \% & +5.3\%\\
    \hline
    White circles & 23 & 19100 & 7200 & 68.6\% & 70.1\% & +1.5\%\\
    \hline 
    Blue rectangles & 6 & 6679 & 2074 & 65.1\% & 66.5\% & +1.4\% \\
    \hline
  \end{tabular}
\end{center}
\end{table*}

\begin{figure*}[t]
\begin{center}
\begin{tikzpicture}
  \begin{axis}[
  	x label style={at={(axis description cs:0.5,-0.1)},anchor=north},
    xlabel=Class of traffic sign,
    ylabel=Absolute difference,
    anchor=south,
    width=18cm,
    height=5cm,
    ybar,
    enlarge x limits=0.02,
    legend style={at={(0.5,-0.2)},
    anchor=north,legend columns=-1},
    bar width=7pt,
   symbolic x coords={A0115, A0130, A0150, A0160, A0170, A0180, A0190, A0430, A0440, A0450, A0460, A0470, A0480, C01, C06, C07, C08, C09, C12, C13, C14, C15, C16, C22, F01, F03, F05, F07, F10, J01, J02, J03, J04, J05, J08, J09, J10, J11, J15, J16, J17, J18, J19, J20, J22, J23, J25, J27, J29, J31, J32, J33, J36, J38, J39},
    xtick=data,
    x tick label style={rotate=90,anchor=east, font=\scriptsize}
    ]
    \addplot [ybar,fill=cyan] coordinates {
    (A0115, 0.3)
    (A0130, -2.1)
    (A0150, 8.8)
    (A0160, 3.0)
    (A0170, 3.8)
    (A0180, 0.6)
    (A0190, -3.5)
    (A0430, 2.8)
	(A0440, 2.0)
	(A0450, 0.9)
	(A0460, 3.4)
	(A0470, -6.6)
	(A0480, -5.1)
    (C01, -0.7)
    (C06, 1.4)
    (C07, 3.1)
    (C08, -1.8)
    (C09, -1.4)
    (C12, 1.1)
    (C13, 5.8)
    (C14, 2.4)
    (C15, 3.6)
    (C16, 1.3)
    (C22, -1.6)
    (F01, 0.2)
    (F03, 3.4)
    (F05, -1.2)
    (F07, 5.7)
    (F10, -8.3)
	(J01, 14.5) 
	(J02, 1.8)
    (J03, 4.5)
    (J04, 4.1)
    (J05, 2.7)
    (J08, 6.2)
    (J09, 7.7)
    (J10, 11.4) 
    (J11, -1.3)
    (J15, -1.8)
    (J16, -3.9)
    (J17, 7.8)
    (J18, 2.1)
    (J19, 4.1)
    (J20, 9.6)
    (J22, 1.6)
    (J23, 20.8)
    (J25, 0) 
    (J27, 8.2)
    (J29, 1.8)
    (J31, 2.5)
    (J32, -0.2) 
    (J33, 5.7)
    (J36, 4.8)
    (J38, 9.7)
    (J39, -4.3)
    };
  \end{axis}
\end{tikzpicture}
\caption{Overview of the classification performance of traffic signs across 55 classes. The $y$ axis represents the absolute difference of top-1 accuracy with respect to the baseline (trained with real data). The $x$ axis represents the class of traffic sign which is described in \cite{ts_dutch}.}
\end{center}
\end{figure*}
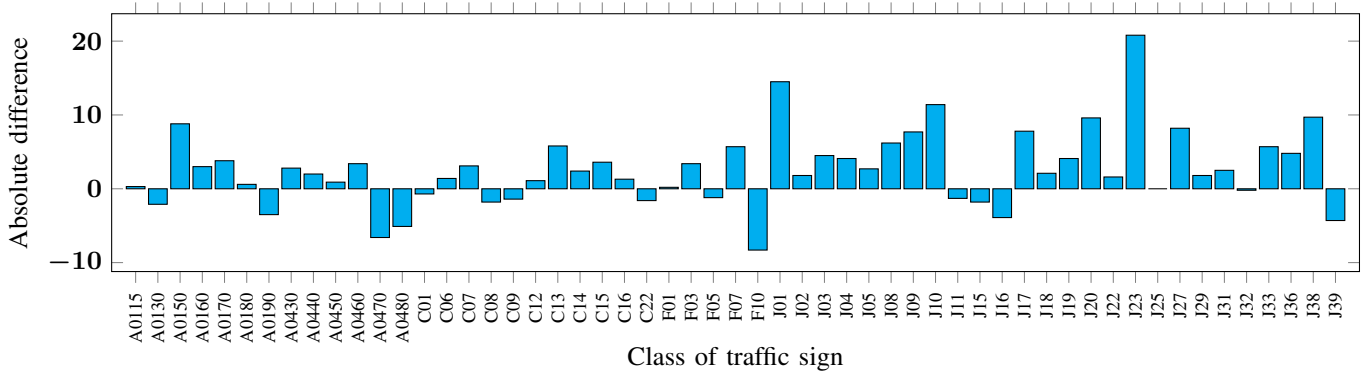

\subsection{Implementation details}
We found that it is easier to transfer traffic signs from a pictogram within a category rather than transferring it across categories. Therefore, we use a model for each category (each category has several classes). For the generator network, a ResNet with encoder-decoder structure is applied as the backbone architecture. The residual connection is replaced by \textit{Dense Residual Attention}, while multi-scale discriminators are applied. The generator up-samples the output at each scale using bilinear up-sampling, followed by convolution with residual units. We use a discriminator at each scale, resulting in a total of three discriminators. The discriminators have networks of varying complexity depending on the image scale. We use three residual units for the image generated at the highest resolution (80~$\times$~80 pixels), two residual units at 40~$\times$~40 pixels and a single residual unit at 20~$\times$~20 pixels. The last layer of each of the discriminator network is a fully-connected layer. Adam optimizer~\cite{DBLP:journals/corr/KingmaB14} is applied to each discriminator. For the mask, we conduct experiments with both rectangular and circular masks across all classes. However, we have found that the circular mask around the desired object offers the best performance, irrespective of the shape of the traffic sign. Depending on the class of traffic sign, we have trained the model between 60K and 100K iterations. Generation of 1000 images took approximately 57 seconds on a NVIDIA Tesla P100 GPU (12 GB).

\subsection{Ablation study}
With a standard ResNet structure for the generator, we observe that the generated traffic sign has a poor geometry and texture. It also produces backgrounds that are unrealistic and have low perceptual quality. Besides this, we note that noise present at the lower scales of the network is up-sampled, resulting in large noisy patches. This is shown in Figure~\ref{fig:fail_cases}. We conduct an ablation study to understand the contribution of each component of the proposed network. The results using our method are illustrated in Figures~\ref{fig:results} and~\ref{fig:results_multi}.

\subsubsection{Dense Residual Attention}  The addition of residual attention mechanisms suppress the noise and produce a clearer background. We have observed a further improvement in the detail of the visual quality after concatenating the encoder feature maps. The dense concatenation partially improves the geometry of the traffic sign. However, the dense concatenation without residual attention does not retain the background information from the input image. 

\subsubsection{Multi-scale discriminators} The multi-scale discriminator captures both the global and local details of the desired image from the input image and pictogram. Without the multi-scale discriminator, the edges and other geometrical features of the traffic sign are not sharp. The outputs at the smaller scale produce consistent geometries which guide the higher-resolution outputs. At the smallest scale, the details of traffic sign texture are absent, whereas the lighting condition and pose are learned. The concatenation of the pictogram at every scale captures the texture of the traffic sign.

\subsubsection{Mask} With the addition of a mask, we can observe an improved performance in replacing the texture of the traffic sign. We also observe a better performance in situations where the traffic sign has a complex pose (skewed angle) or texture. 

\subsection{Detection and classification results using generated data}
We have generated samples using our method and added the new data to the existing training set. For training, we use an ensemble of HOG-SVM and CNN detectors \cite{dalal2005histograms}\cite{creusen2010color}. For detection, the baseline recall is ~99\% on the test set and we did not notice any significant improvement in the recall with the addition of generated sets. However, the number of false positives decreased between 1.2\% - 1.5\% in the detection tests. This is beneficial in reducing manual efforts to remove false-positives from the automatically extracted detections. We have also conducted classification tests using the generated data. The details of the classification results are shown in Table 1. Out of the 313 classes, 55 classes that have a low amount of data, are expanded. Among the 55 classes, 41 classes produced a higher classification rate when trained with the combination of real and generated samples. The lowest score obtained for a class is 8.3\% (average decrement by 2.92\%) below the baseline, whereas the highest gain has 20.8\% absolute improvement (average increment by 4.65\%) over the baseline. We do not conduct detection and classification performance analysis of classes outside the training set as there is no real test set.

\subsection{Failure cases and comparison with other methods}

\begin{figure}[h]
\begin{center}
   \includegraphics[width=1.0\linewidth]{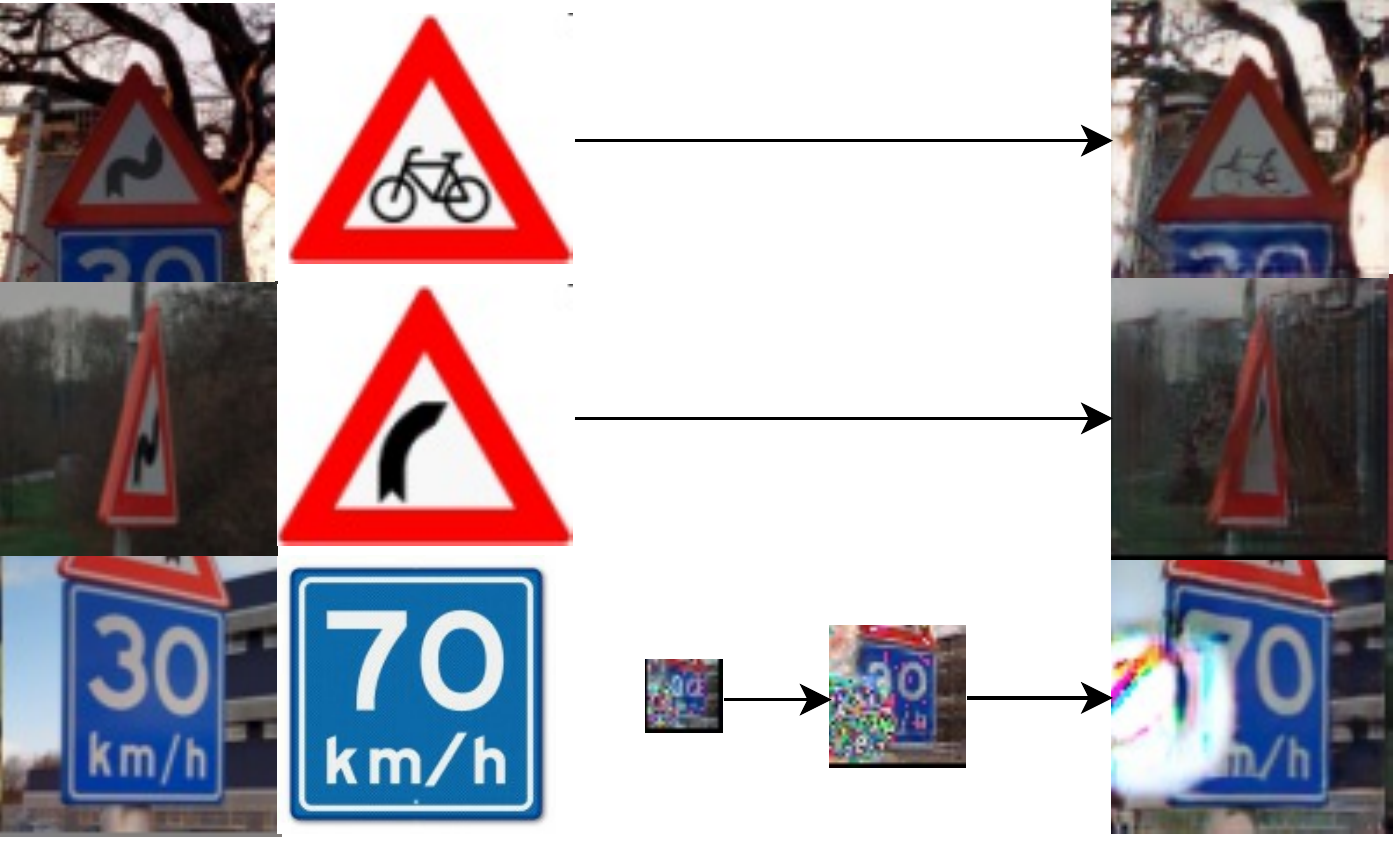}
\end{center}
   \caption{ Examples of failure using our method. Top row: Generated output lacks details in the traffic sign and background. Middle row: Skewed angled traffic signs have difficulty producing pictogram textures. Bottom row: Noise present in the lower layers progressively grow into a large noise. }
\label{fig:fail_cases}
\end{figure}

Figure \ref{fig:fail_cases} demonstrates examples of failure cases with our method. In the first and second row, we observe lack of details when the traffic sign in the pictogram are not thick or when  traffic signs have a skewed angle.
At the bottom row (Figure \ref{fig:fail_cases}), we observe noisy outputs that progressively become larger, as the network scale increases in the generator. However, the residual attention mechanism exhibits a certain amount of robustness to this type of noise. We observed fewer cases with such noise compared to Conditional GANs.  

\begin{figure}[htbp!]
\begin{center}
   \includegraphics[width=1.0\linewidth]{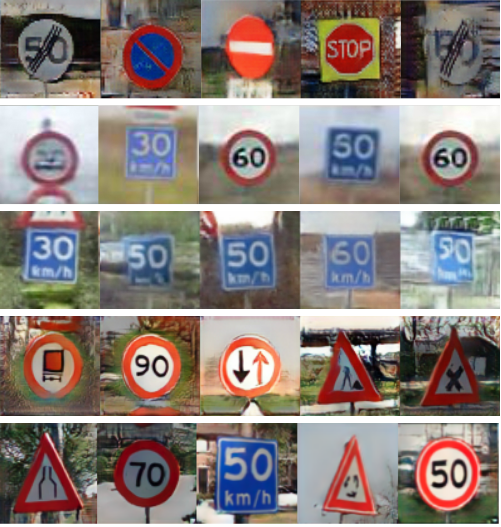}
\end{center}
   \caption{Row 1 (top): WGAN-GP (generated from noise), Row 2: Boundary Equilibrium GAN (generated from noise), Row 3: Conditional Analogy GAN, Row 4: Conditional GAN, Row 5 (bottom): Conditional GAN with Dense Residual Attention and multi-scale discriminators (our method). }
\label{fig:wgan}
\end{figure}

We conducted experiments with other methods as well and results are presented in Figure \ref{fig:wgan}. WGAN-GP generates images from noise, which results in smeared backgrounds and traffic signs. Boundary Equilibrium GAN (BEGAN) often results in mode collapse (image generated from noise) that produce a low variety of samples. Conditional Analogy GAN (CAGAN) uses an implicit mask generation fails with street-view images due to complex backgrounds which result in poor outputs. Conditional GANs (cGANs) produces outputs, which are often smudged and have the incorrect geometry of the traffic sign. 

\section{Conclusions}
We have presented a conditional GAN with Dense Residual Attention, which generates a new traffic sign conditioned on a given pictogram. The network utilizes multiple Dense Residual Attention modules that are composed of a residual attention mechanism and a dense concatenation. The Dense Residual Attention module improves the visual quality of the traffic signs and suppresses the cases with progressively growing noise. We propose the use of multi-scale discriminators, which result in images of traffic signs that are both globally and locally coherent. The discriminator at smaller scale captures global features and steers the high-resolution output to produce images with more accurate geometries. The discriminator at the larger scale along with the concatenated pictogram assists in producing images of traffic signs with finer details. Comparison other methods reveals that the proposed method produces visually appealing results with finer details in the traffic signs and has fewer geometrical errors. We have further conducted detection and classification tests across a large number of traffic sign classes, by training our detector with the combination of real and generated data. The trained model reduces the number of false positives by about 1.2 - 1.5\% at a recall of 99\% in the detection tests while improving the top-1 accuracy on the average by 4.65\% in classification tests.






%
{\small
\bibliographystyle{IEEEtran}
\bibliography{ref}
}

\end{document}